\documentclass[twoside,leqno,twocolumn]{article}
\usepackage[letterpaper]{geometry}
\usepackage{makecell}
\usepackage{graphicx}
\usepackage{subfig}
\usepackage{caption}
\usepackage{float} 
\usepackage{adjustbox}
\usepackage{ltexpprt}
\usepackage{hyperref}
\usepackage{amsmath,amsfonts}
\usepackage{mathrsfs}
\usepackage{url}

\usepackage{amssymb}
\usepackage[justification=centering]{caption}
\usepackage{soul}
\usepackage{xcolor}
\usepackage{booktabs}
\usepackage{multirow}
\begin{document}
\newcommand\relatedversion{}
\newcommand\blfootnote[1]{%
  \begingroup
  \renewcommand\thefootnote{}\footnote{#1}%
  \addtocounter{footnote}{-1}%
  \endgroup
}
\newtheorem{definition}{Definition}
\title{\textbf{NEEDED: Introducing Hierarchical Transformer to Eye Diseases Diagnosis}}
\author{Xu Ye$^{1,2,\S}$ \and Meng Xiao$^{1,2,\S}$ \and Zhiyuan Ning$^{1,2}$ \and Weiwei Dai$^3$ \and Wenjuan Cui$^1$ \and Yi Du$^{1,*}$  \and Yuanchun Zhou$^1$}
\date{} 
\maketitle
\let\thefootnote\relax\footnotetext{{\scriptsize $^1$ Computer Network Information Center, CAS, Emails:\\ yexu@cnic.cn, shaow@cnic.cn, ningzhiyuan@cnic.cn, wenjuancui@cnic.cn, duyi@cnic.cn, zyc@cnic.cn}}
\let\thefootnote\relax\footnotetext{\scriptsize $^2$ University of Chinese Academy of Sciences}
\let\thefootnote\relax\footnotetext{\scriptsize $^3$ Changsha Aier Eye Hospital, Emails:	daiweiwei@aierchina.com}
\let\thefootnote\relax\footnotetext{\scriptsize $\S$ These authors have contributed equally to this work.}
\let\thefootnote\relax\footnotetext{\scriptsize $^*$  Corresponding author}

\fancyfoot[R]{\scriptsize{Copyright \textcopyright\ 2023 by SIAM\\
Unauthorized reproduction of this article is prohibited}}

\begin{abstract}
\small\baselineskip=9pt 
With the development of natural language processing techniques(NLP), automatic diagnosis of eye diseases using ophthalmology electronic medical records (OEMR) has become possible. 
It aims to evaluate the condition of both eyes of a patient respectively, and we formulate it as a particular multi-label classification task in this paper.
Although there are a few related studies in other diseases, automatic diagnosis of eye diseases exhibits unique characteristics.
First, descriptions of both eyes are mixed up in OEMR documents, with both free text and templated asymptomatic descriptions, resulting in sparsity and clutter of information. 
Second, OEMR documents contain multiple parts of descriptions and have long document lengths.
Third, it is critical to provide explainability to the disease diagnosis model. 
To overcome those challenges, we present an effective automatic eye disease diagnosis framework, NEEDED.
In this framework, a preprocessing module is integrated to improve the density and quality of information. 
Then, we design a hierarchical transformer structure for learning the contextualized representations of each sentence in the OEMR document.
For the diagnosis part, we propose an attention-based predictor that enables traceable diagnosis by obtaining disease-specific information.
Experiments on the real dataset and comparison with several baseline models show the advantage and explainability of our framework.  \let\thefootnote\relax\footnotetext{the code can be found in \url{https://github.com/coco11563/NEEDED}}

\end{abstract}




\section{Introduction}
With the rising incidence of eye diseases and the shortage of ophthalmic medical resources, eye health issues are becoming increasingly prominent~\cite{whoreport}.
To improve the efficiency of ophthalmologists and lower the rate of misdiagnosis, an automated eye disease diagnostic system that provides advice to ophthalmologists has become an urgent demand.
There have been many attempts to automatically diagnose eye diseases based on image data (e.g., ocular fundus photographs)~\cite{christopher2018performance,keel2019visualizing,milea2020artificial}. However, a kind of image usually contains information about only one part of the eye, limiting the scope of automatic disease diagnosis.
Meanwhile, the widespread adoption of ophthalmology electronic medical record (OEMR) and the development of NLP techniques make it possible to diagnose eye diseases based on OEMR automatically.

As shown in the left part of Figure \ref{fig:framework}, the OEMR document consists of several parts, including Chief Complaint (CC), History of Present Illness (HPI), Examination Results (ER), etc. Each part can be considered a particular type of text.
The goal of automatic diagnosis of eye diseases is to diagnose diseases suffered by each of the patient's eyes respectively based on these texts.
For this task, a typical solution that has been tried on similar tasks is to concatenate these texts and use neural networks like RNN~\cite{hashirMortality,sha2017rnn,schoene2019rnn} and CNN~\cite{yang2018cnn,girardi2018cnn} to obtain the embedding representation of the OEMR document. Then the document-level representation is used as input to classifiers to obtain results.

However, after analyzing the characteristics of OEMR and our task, we found three issues that make this solution unsuitable: \textbf{Issues 1: sparsity and clutter of information in the OEMR}.
Descriptions of different eyes are mixed up in OEMR documents, causing clutter in the information relevant to different eyes. Besides, many of these descriptions are templated asymptomatic descriptions such as \textit{No congestion or edema in the conjunctiva of the left eye}, resulting in a sparsity of information in the document. This sparsity and clutter of information may hinder the diagnosis model from giving the accurate diagnosis results of two eyes respectively.
\textbf{Issues 2: long document length and multiple text types}. OEMR documents are long and have multiple text types. This solution is inappropriate when the input is a lengthy document because it is weak at learning long-range dependencies, making it challenging to extract information from long documents comprehensively~\cite{vaswani2017attention,xiao2021expert}. Besides, it ignores the differences between text types and does not preserve the type information of the text.
\textbf{Issues 3: explainability of automatic disease diagnosis}. It is critical to provide explainability to the proposed model for medical scenarios, especially for automatic disease diagnosis~\cite{zhang2022applications}. However, This typical approach obtains only a coarse-grained document-level embedding representation to diagnose all diseases, making it difficult to trace the diagnosis and explain the results.

Another method used in related tasks is extracting structured information from clinical texts by entity recognition and other methods, then using them to enhance the downstream model or get the diagnosis results directly~\cite{chen2020towards,yuan2021graph}. 
This method requires annotated dataset to train the information extraction model. However, due to the complexity and length of OEMR documents, doctors need to spend too long time annotating comprehensively and accurately, making this method difficult to implement. Besides, it suffers from the problem of error propagation.

To overcome these issues, we formulate the automatic eye disease diagnosis as a particular multi-label classification task. Specifically, the input to this task is an OEMR document, and the output is two sets of labels corresponding to the diagnosis results for both eyes, respectively. Then, we i\textbf{N}troduce hierarchical transformer to \textbf{E}y\textbf{E} \textbf{D}is\textbf{E}ases \textbf{D}iagnosis and propose an automatic diagnosis framework named NEEDED.
The main contributions of this paper are summarized as:

\smallskip
\noindent$\bullet$ \textbf{An efficient preprocessing method for improving information density and quality.}
We first filter out useless asymptomatic descriptions from the OEMR document and then distinguish the contents relevant to different eyes to form two documents, which will be used separately to get the diagnostic result.
While retaining vital information, we improve the information density and quality of the document.

\smallskip
\noindent$\bullet$ \textbf{A hierarchical encoder to extract abundant semantic information from OEMR.}
Inspired by the long-text modeling methods~\cite{pappagari2019hierarchical,zhang2019hibert,liu2019hierarchical}, we design a hierarchical encoder architecture to embed both texts and their particular type within an OEMR document into a matrix.
Specifically, we utilize a hierarchical transformer~\cite{xiao2022who,xiao2022hierarchical} to obtain the contextualized representation of each sentence in the OEMR while preserving the type information of the sentence by adding type-token~\cite{xiao2021expert}. Then, we use the matrix formed by these representations as the document representation.

\smallskip
\noindent$\bullet$ \textbf{An attention-based predictor that enables traceable diagnosis.}
We utilize a dot-product attention layer as an extractor to capture disease-specific information from these sentence representations and perform the diagnosis. This method enables traceable diagnosis by observing the distribution of attention weight and providing explanations for diagnostic results.

\smallskip
We conduct experiments on the real dataset and comparison with several baseline models to validate the advantage of NEEDED and provide a case study to show the explainability.

\section{Preliminaries}
In this section, we first give the definition of the ophthalmology electronic medical record(OEMR), then provide a formal definition of the task of Automatic Diagnose of Eye Disease(ADED).

\begin{figure*}[!ht]
\centering
\includegraphics[width=1.0\linewidth]{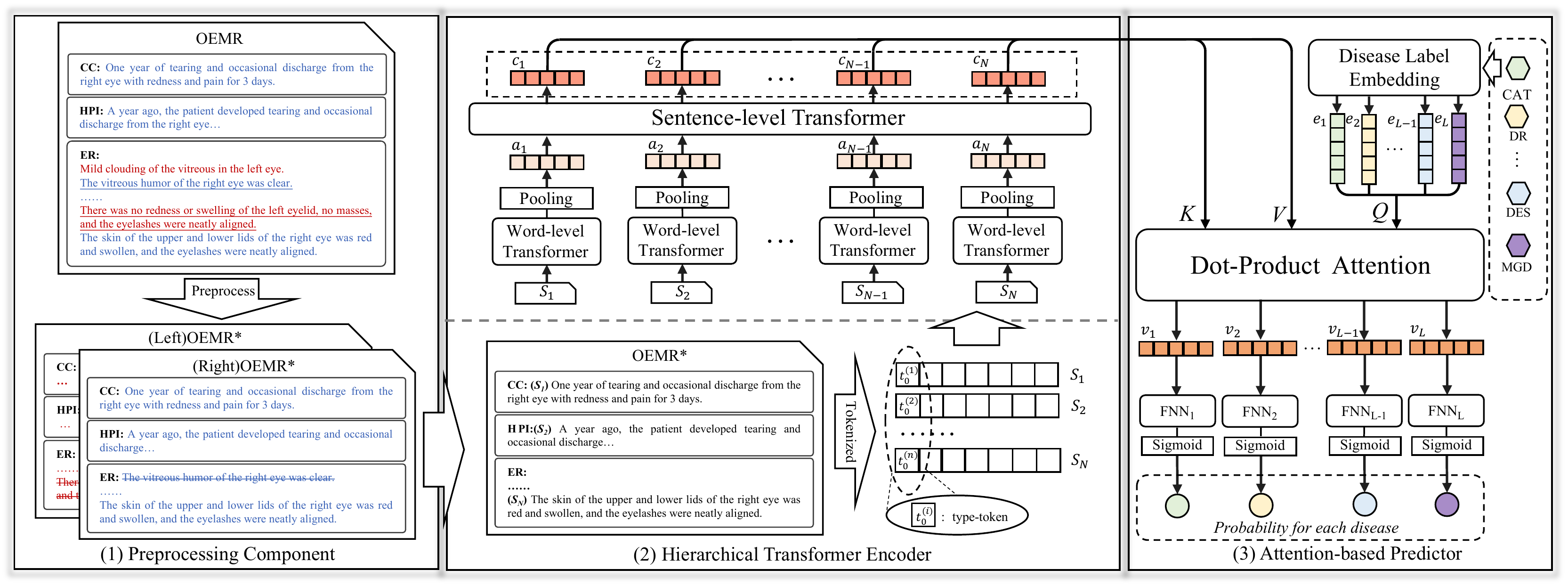}
\caption{The NEEDED takes a given OEMR $O$ as input and outputs two disease label sets corresponding to diffrent eyes of the patient. (1) Preprocessing Component.
The useless parts of the descriptions marked by underline are first filtered out, and then the descriptions relevant to different eyes are distinguished to form two documents, $O_l$ and $O_r$. They will be used separately to get the diagnostic results for the two eyes. (2) Hierarchical Transformer Encoder. Taking $O_r$ for example, each sentence set $\left\{S_1, S_2, \dots, S_N\right\}$ of $O_r$ are encoded into the document representation $C_r$. (3) Attention-based predictor. The predictor captures disease-specific information representation from $C_r$ and outputs the disease's presence probability.}
\label{fig:framework}
\vspace{-0.5cm}
\end{figure*}

\noindent\textit{Definiton 1 \textbf{(Ophthalmology Electronic Medical Record):}} As shown in the left part of Figure \ref{fig:framework}, an OEMR has several parts. Each part can be considered as a set of sentences. Therefore, the OEMR can be viewed as a document consisting of multiple sentences. Formally, we define $O = \{S_1, S_2, ..., S_N\}$ as an OEMR document with N sentences. The $i$-th sentence in the document is defined as $S_{i} = [t_1^{(i)} , t_2^{(i)} , ..., t_{|S_i|}^{(i)} ]$, where $t_k^{(i)}$ denotes the $k$-th token in $i$-th sentence.

\noindent\textit{Definiton 2 \textbf{(Automatic Diagnosis of Eye Disease):}} Given an OEMR $O$, the task of automatic diagnosis of eye diseases aims to diagnose the diseases suffered by each of the patient's eyes respectively based on it.
By treating each disease as a label, we can view this problem as deriving two sets of labels for each OEMR corresponding to different eyes, which can be seen as a particular multi-label classification task. 
Formally, Let $L = \{l_1, l_2, ..., l_D\}$ denotes the label set for all eye disease, where $l_i \in \{0, 1\} $ represent the presence or absence of the $i_{th}$ disease. As mentioned before, we consider that it is essential to preprocess the OEMR document, let $\digamma$ denote the preprocessing method. Finally, the ADED task can be formulated as:
\begin{equation}
\Omega(\digamma(O), \Theta) \rightarrow L_{a}, L_{b}
\end{equation}
where $\Theta$ is the parameters of the disease diagnosis model $\Omega$, $L_{a}$ and $L_{b}$ denote the set of diseases labels for two eyes respectively.


\section{Proposed Framework}


\subsection{Preprocessing Component} 
As shown in the left part of Figure \ref{fig:framework}, the OEMR document is composed of several sections, including Chief Complaint (CC), History of Present Illness (HPI), Examination Results (ER), etc. Each part can be considered a particular type of text that consists of some sentences. The task of automatic eye disease diagnosis aims to evaluate the status of both eyes respectively based on an OEMR document. However, descriptions of different eyes are mixed up in OEMR documents, resulting in a clutter of information.
Besides, many descriptions are templated asymptomatic descriptions that contain little information for disease diagnosis, their abundance results in a sparsity of information in the document.

To improve information density and quality, we propose a preprocessing method consisting of two steps: filtering useless descriptions and distinguishing content relevant to different eyes.
\ul{\textit{Step-1:}} Inspired by TF-IDF, we consider that for those templated asymptomatic descriptions, the more they appear in the document set, the more they are common condition descriptions and the less useful they are for disease diagnosis. Thus, we calculate the frequency of occurrence of each description in all OEMR documents and filter out asymptomatic descriptions that occur more frequently than the threshold $B$.
\ul{\textit{Step-2:}} Given an OEMR document $O$, the descriptions relevant to different eyes are distinguished from $O$ to form two documents, $O_l$ and $O_r$. They have the same structure as the OEMR, except that the content is focused on a particular eye. Specifically, for the examination results section, each sentence describes an examination result for a specific eye, so we can easily distinguish them. For other parts, we developed some rules based on the characteristics of the descriptions to distinguish, and for those sentences that are difficult to distinguish or relevant to both eyes, both documents retain them.

\subsection{Hierarchical Transformer Encoder} 
After the above stage, we obtain two documents, $O_l$ and $O_r$. They will be used as input of the downstream model separately to get the diagnostic results for the two eyes. For simplicity of expression, they will be denoted as $O_{\ast}$ in the following.
To extract the information in a given OEMR document $O_{\ast}$, we design a hierarchical Transformer structure as an encoder.
As shown in Figure \ref{fig:framework}(2), the encoder consists of a word-level Transformer and a sentence-level Transformer. The word-level transformer takes the OEMR document as input and encodes each sentence in the document to obtain the type-specific sentence embedding representations. 
Then, the sentence-level transformer takes the set of sentence embedding representations as input, learns the dependencies between sentences and generate the representation matrix of the input OEMR document. 


\subsubsection{Word-level Transformer}
Given a OEMR document $O_{\ast}$, in this stage we aims to use a Transformer to embed each sentence $S_i$ in $O_{\ast}$ into the sentence representation $a_i \in \mathbb{R}^d$. To retain information about the text type of each sentence and thus help the Transformer model its semantics, we first add a type-token $t_0$ at the beginning of each sentence. Formally, given an sentence $S_i$ with token sequence $[t_0^{(i)}, t_1^{(i)}, t_2^{(i)}, \dots, t_{|S_i|}^{(i)}]$, the sentence representation $a_i$ can be obtained by: 

\begin{equation}
\begin{aligned}
 a_i &= Pooling([h_0^{(i)}, h_1^{(i)}, h_2^{(i)}, \dots, h_{|S_i|}^{(i)}]]),\\
     &= Pooling((Transformer_{w}([t_0^{(i)}, t_1^{(i)}, t_2^{(i)}, \dots, t_{|S_i|}^{(i)}])),
\end{aligned}
\end{equation}
where $Transformer_{w}$ is a $N_{w}$-layers Transformer. For each token  $t_j^{(i)}$ in the sentence $S_i$, its input embedding is obtained by summing its token embedding and positional encoding, and the token embedding is randomly initialized.
$[h_0^{(i)}, h_1^{(i)}, h_2^{(i)}, \dots, h_{|S_i|}^{(i)}]$ is a list of output embedding of tokens in the sentence $S_i$, where $h_j^{(i)}$ is a $d$ dimensional vector corresponding to $t_j^{(i)}$. 
The $pooling(\cdot)$ denotes the average-pooling on the output embedding representations to obtain the sentence embedding representation $a_i$.

\subsubsection{Setence-level Transformer}
In the above stage, we obtain the representation of each sentence in the OEMR document. Most of the sentences are descriptions of the patient's symptoms or medical history. It is likely that there are many associations between them, which means that the meaning represented by one sentence may vary depending on its context. Thus, the embedding representation of each sentence should not only depend on its content but also consider its context. 

At this stage, we aims to use the Transformer to obtain contextualized sentence representations by learning the dependencies between sentences. Given a OEMR document $O_{\ast}$, we can obtain the sentences representations set $\left\{a_1, a_2, ..., a_N\right\}$ from the above word-level Transformer.
The setence-level transformer takes sentences representations set as the input and outputs a matrix $C$, which can be formulated as:

\begin{equation}
\begin{aligned}
 C &= [c_1,c_2, \dots, c_N] \\
     &= Transformer_s\left([a_1, a_2, \dots , a_N]\right)   
\end{aligned}
\end{equation}
where $Transformer_{s}$ is a one layer Transformer. $C = [c_1,c_2, \dots, c_N]$ is a set of contextualized sentence representations. With the multi-head self-attention mechanism, Transformer can model the dependencies between sentence representations in the input from multiple perspectives, enabling each sentence representation can collect global information of context. In a sense, the matrix $C$ can be seen as a multi-perspective representation of the input OEMR document.

The hierarchical transformer encoder has the following main advantages: 
First, Transformer has strengths in learning long-range dependencies~\cite{vaswani2017attention,xiao2021expert}. The hierarchical Transformer encoder can effectively and efficiently model the semantic information of OEMR by learning both word-level and sentence-level dependencies.
Second, embedding documents into multiple sentence-level representations allows us to implement traceable disease prediction based on these representations in combination with our attention-based predictor proposed below.
Third, by adding the type-token at the beginning of sentences, the differences between text types are taken into account, helping the encoder to model sentence semantics more accurately.
 
\subsection{Attention-based Predictor}
After the previous step, the OEMR document is embedded in a matrix $C \in \mathbb{R} ^{d  \times N} $, where $N$ denotes the number of sentences in the OEMR document. A typical subsequent operation is to perform a pooling operation on matrix $C$ to obtain a $d$ dimensional vector and use this vector as the input to classifiers to obtain the diagnostic result. However, this approach obtains only a coarse-grained document-level representation to diagnose all diseases, making it difficult to trace the diagnosis and provide explanations for the results.

To provide explainability for the automatic diagnosis, inspired by the label-wise attention\cite{CAML}, we adopt the dot-product attention to capture disease-specific information from $C$ for traceable diagnosing. 
Specially, given the disease $l$ to be predicted, we first obtain the attention weight of each component in $C$ for disease $l$. this process can be formulated as:
\vspace{-0.1cm}
\begin{equation}
\alpha_{l}=\operatorname{softmax}\left(\frac{C^{\top}e_{l}}{\sqrt{d}}\right)
\end{equation}
\vspace{-0.1cm}
where the $e_l \in \mathbb{R}^{d}$ is the label embedding of disease $l$, which is randomly initialized. The $\alpha_{l} \in \mathbb{R}^{d}$ is the attention weight vector for disease $l$.
Then, we use the attention weight vector $\alpha_{l}$ and matrix $C$ to compute the specific representation $v_l$ for disease $l$.  
\begin{equation}
v_{l} = C \alpha_{l} = \sum_{n=1}^{N} \alpha_{l}^{n} c_{n}
\end{equation}

Be noted that each component in $C$ is an embedding representation of a sentence in the input. Thus, by observing the attention weight vector $\alpha_{l}$, we can know the importance of sentences in the input for diagnosing disease $l$, thus enabling a traceable diagnosis.

Suppose we have a total of $L$ diseases to diagnose, after the above stage, we get a matrix $V = [v_1,v_2, \dots, v_L]$ where $v_{l}$ is the disease-specific information representation for disease $l$.
Then, we use a set of Feedforward Neural Networks(FNNs) and the sigmoid function to calculate the probability of the presence of each disease.
\begin{equation}
\hat{y}_{l} = \operatorname{sigmoid}(FNN_{l}(v_{l}))
\end{equation}

\subsection{Training}
The training procedure minimizes the binary cross-entropy loss, which can be formulated as:
\begin{equation}
L=-\sum_{l=1}^{L} y_{l} \log (\hat{y}_{l}) + (1 - y_{l}) \log (1 - \hat{y}_{l})
\end{equation}

where $y_{l} \in \left\{0,1\right\}$ indicates whether disease $l$ exists in the diagnosis record of the input document. $\hat{y}_{l}$ is the probability of the existence of disease $l$ output by the prediction model.


\section{Experiments}
This section presents our experiments' dataset and the evaluation of experimental results with some analysis. We demonstrate the advantage of our proposed framework by comparing it to the baseline models. We explore the effectiveness of each component of our framework and provide a case study to illustrate the explainability.

\subsection{Dataset Description}
We conducted the experiments on the real-world OEMR dataset with 5134 records collected by Aier Eye Hospital. 
After discussion with ophthalmologists,
we select six common eye diseases to test the diagnostic ability of our model, including cataract(CAT), glaucoma, diabetic retinopathy(DR), dry eye syndrome(DES), pterygium, and meibomian gland dysfunction(MGD). The data set is divided into a training set, validation set, and test set according to the proportion of 70\%, 15\%, 15\%, respectively. 

\subsection{Baseline Methods}
To comprehensively evaluate the performances of our proposed framework, we selected three baselines and five ablation variants of NEEDED. The baselines are listed as follows: 1) \textbf{CAML~\cite{CAML}:}  The CAML is a CNN-based model which integrate label-wise attention to obtain the label-specific representations.
2) \textbf{LSTM-Att~\cite{schoene2019rnn}:}
LSTM-Att combines LSTM and attention mechanism to capture important semantic information in the input for prediction.
3) \textbf{BERT~\cite{devlin2018bert}:}BERT is the representative of the pre-trained models, we use BERT and an pooling layer to obtain the representation of input.
We also proposed five variants of NEEDED which are listed as follows:
1) \textbf{w/o p}: ablated the pre-processing steps.
2) \textbf{w/o c}: ablated the hierarchical transformers and replaced it with vanilla transformer.
3) \textbf{w/o s}: ablated the sentence-level transformers.
4) \textbf{w/o l}: ablated the attention mechanism and label embedding in predictor.
5) \textbf{w/o w}: both ablated the hierarchical transformers and the attention based predictor. 

\subsection{Experiment Settings}
In experiments, we set the transformer layer number $N_{w}$ to 5, the multi-head number to 8, and the dimension size $d_{model}$ to 256. For the detail of NEEDED training, we use AdamW \cite{adamw} optimizer with learning rate of $1 \times 10^{-4}$, and set the batch size as 32, AdamW weight decay as $1 \times 10^{-2}$. The dropout rate is set to 0.1 to prevent overfitting. For all models, we train and test them multiple times with different random seeds under their optimal hyperparameters and report their performance and standard deviation. In the following experiments, all methods are implemented by PyTorch, and all experiments are conducted on a Linux server with an AMD EPYC 7742 CPU and one NVIDIA A100 GPU.
\vspace{-0.2cm}
\begin{table}[!ht]
\renewcommand\arraystretch{1}
\centering
\caption{The overall comparison experiments. For a fair comparison, we equipped three baselines with the preprocessing and marked them with ${(\cdot)}^\rho$. The best performance is marked by \textbf{bold}, and the runner up is marked by \ul{underlines}.}\label{experiment}
\vspace{-0.2cm}
\setlength{\tabcolsep}{1.0mm}{\resizebox{\linewidth}{!}{
\begin{tabular}{ccccc}
\toprule
Models & {Macro-F1} & {Micro-F1} & {Macro-AUC} & \multicolumn{1}{c}{Micro-AUC} \\ \midrule

BERT  & 81.21$_{\pm0.40}$        & 84.28$_{\pm0.78}$      & 92.55$_{\pm0.43}$        & 95.61$_{\pm0.22}$  \\ 
CAML     & 83.71$_{\pm0.57}$        & 81.22$_{\pm0.80}$                        & 94.32$_{\pm0.61}$                        & 96.09$_{\pm0.20}$      \\ 
LSTM-Att & 81.55$_{\pm0.74}$       & 80.29$_{\pm0.70}$                       & 93.50$_{\pm0.39}$                 & 95.68$_{\pm0.23}$      \\
BERT${^\rho}$  & 87.39$_{\pm0.72}$      & 89.73$_{\pm0.34}$                & 95.73$_{\pm0.33}$            & 97.33$_{\pm0.19}$       \\ 
CAML${^\rho}$       & 88.49$_{\pm0.42}$   & 90.53$_{\pm0.33}$  & 96.10$_{\pm0.18}$   & 97.59$_{\pm0.14}$     \\ 
LSTM-Att${^\rho}$ & 88.19$_{\pm0.61}$                       & 91.11$_{\pm0.44}$      & \underline{97.16}$_{\pm0.23}$          & \underline{98.14}$_{\pm0.12}$        \\\midrule
w/o p        & 82.59$_{\pm0.76}$                        & 82.37$_{\pm0.78}$                        & 94.85$_{\pm0.10}$                         & 96.46$_{\pm0.09}$      \\
w/o c    & 87.96$_{\pm0.73}$     & 89.80$_{\pm0.44}$      & 97.14$_{\pm0.26}$     &   98.00$_{\pm0.19}$   \\
w/o s     &  88.19$_{\pm0.50}$         & 91.31$_{\pm0.34}$           & 96.85$_{\pm0.15}$                  & 97.89$_{\pm0.08}$     \\  
w/o l     &  \underline{89.29}$_{\pm0.65}$     & \underline{91.88}$_{\pm0.49}$        &   96.40$_{\pm0.19}$     & 97.67$_{\pm0.20}$  \\ 
w/o w    & 87.29$_{\pm0.59}$      & 90.58$_{\pm0.36}$      & 96.13$_{\pm0.22}$     &   97.51$_{\pm0.17}$    \\
\midrule
NEEDED        & \textbf{90.25$_{\pm0.48}$}                        & \textbf{92.61$_{\pm0.27}$}                        & \textbf{97.92$_{\pm0.15}$}                         & \textbf{98.59$_{\pm0.10}$}   \\\bottomrule
\end{tabular}
}}
\vspace{-0.4cm}
\end{table}

\subsection{Experiment Results}
These experiment aims to answer the following questions: \textbf{Q1}:How is the performance of NEEDED in compared with other baseline methods? \textbf{Q2:} How is each component of NEEDED impact the model performance? \textbf{Q3:} What is the impact of distinguishing the descriptions of different eyes in preprocessing? \textbf{Q4:} What is the impact of filtering the asymptomatic templated descriptions? \textbf{Q5:} How is the explainablity of our proposed model? Beside this, we also conduct the hyperparameter senetivity study.

\begin{figure*}[!ht]
	\centering
	\subfloat[Micro-F1]{  
		\includegraphics[width=4.0cm]{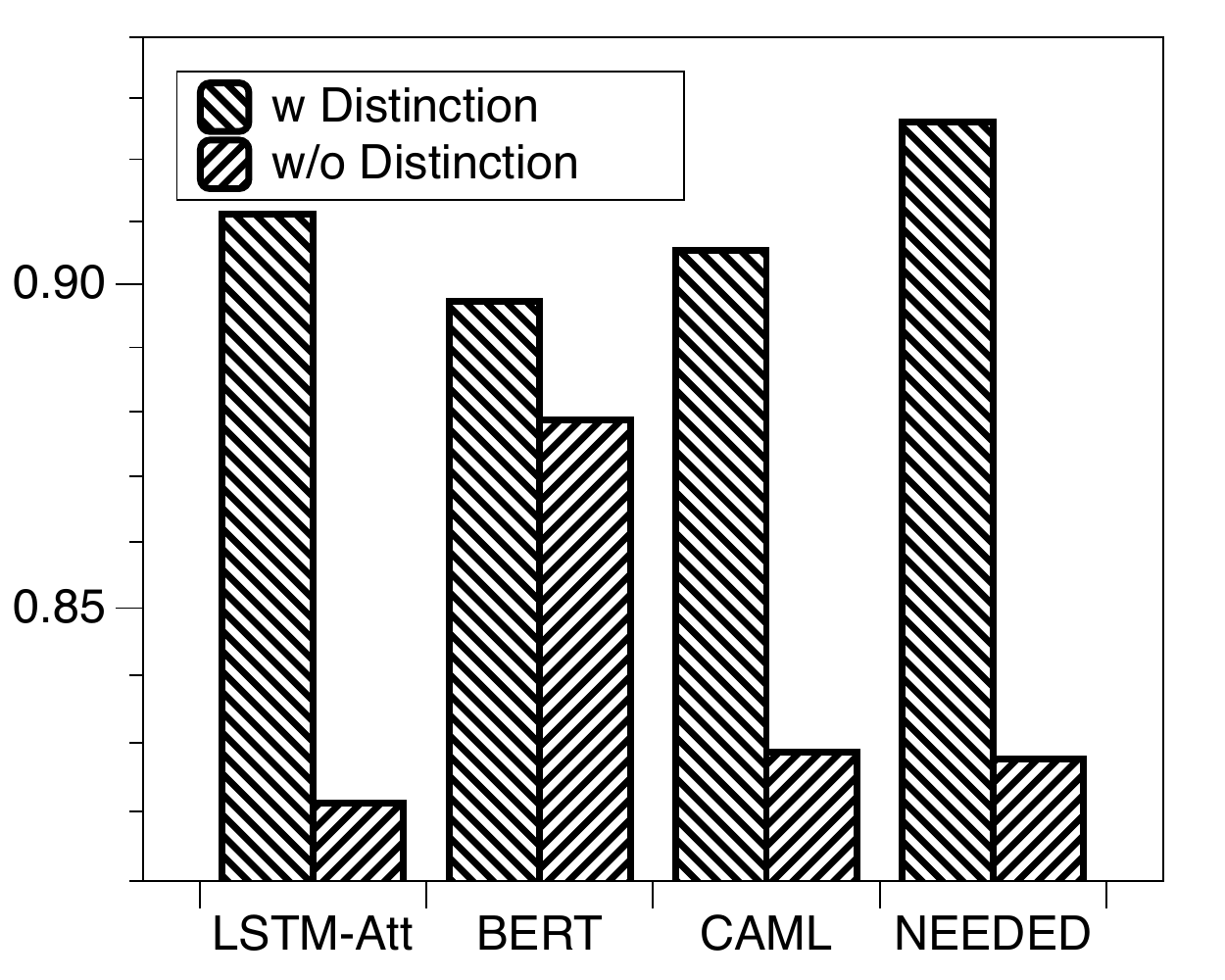}}
		\hspace{-3mm}
	\subfloat[Macro-F1]{
		\includegraphics[width=4.0cm]{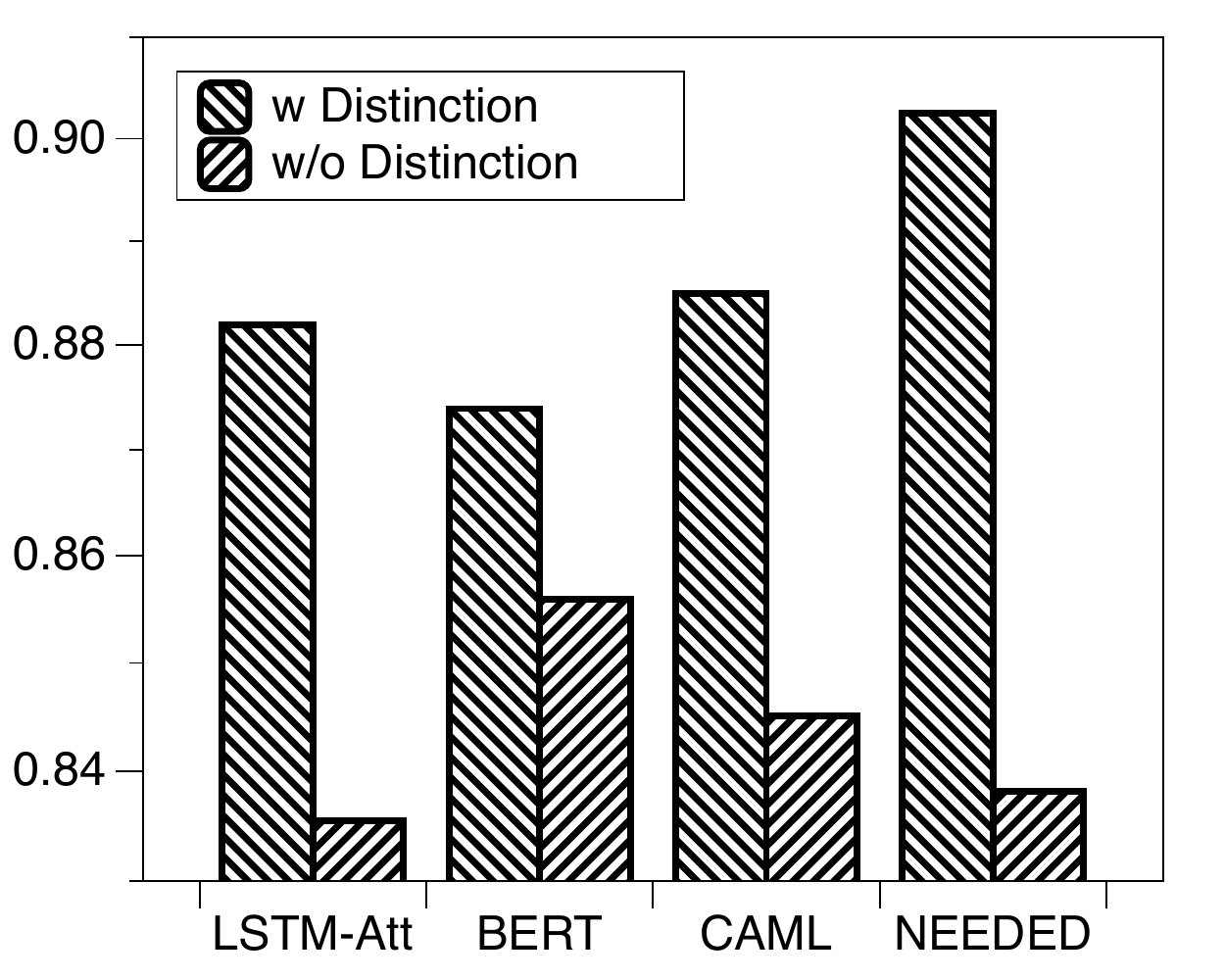}}
		\hspace{-3mm}
	\subfloat[Micro-AUC]{
		\includegraphics[width=4.0cm]{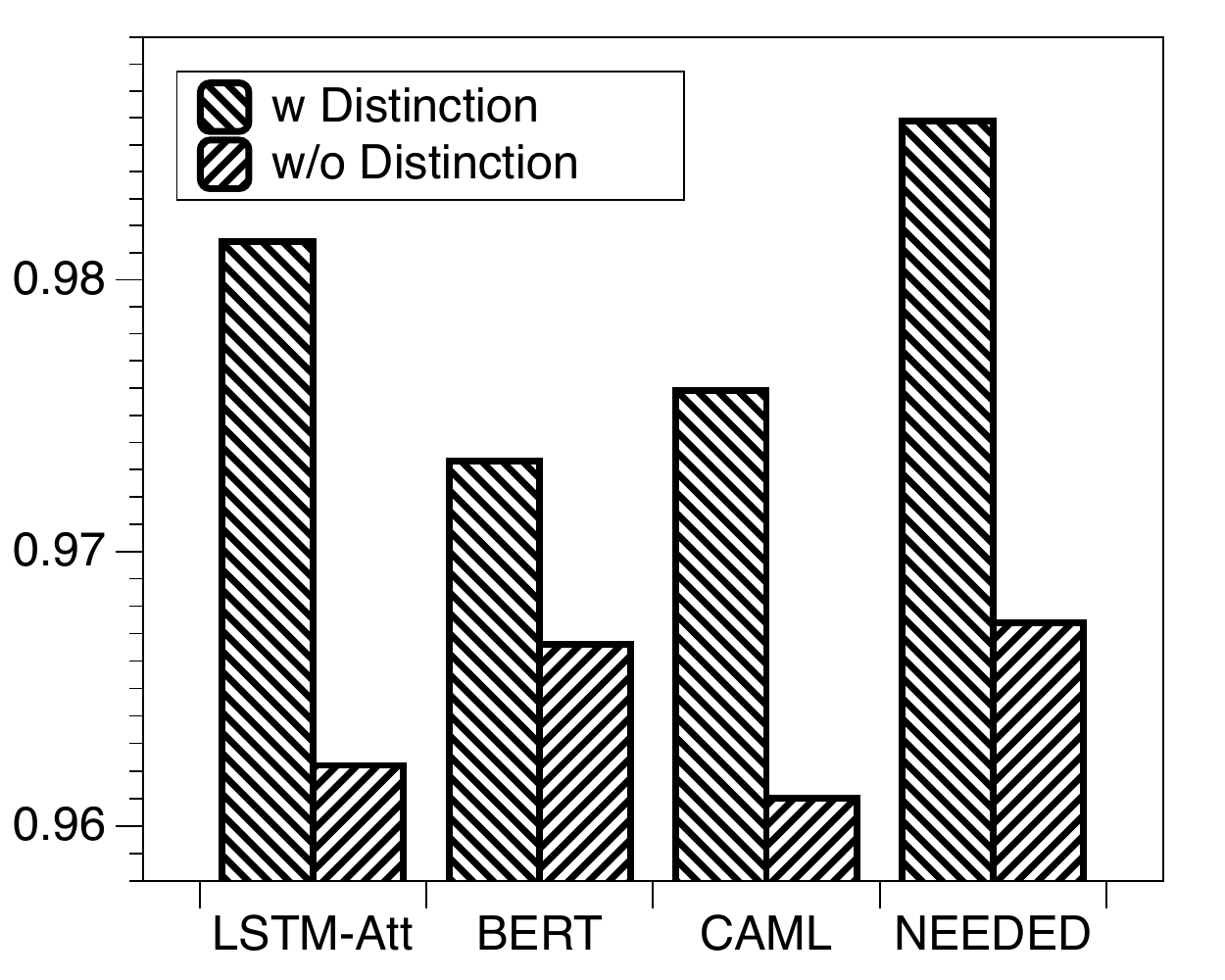}}
		\hspace{-3mm}
	\subfloat[Macro-AUC]{
		\includegraphics[width=4.0cm]{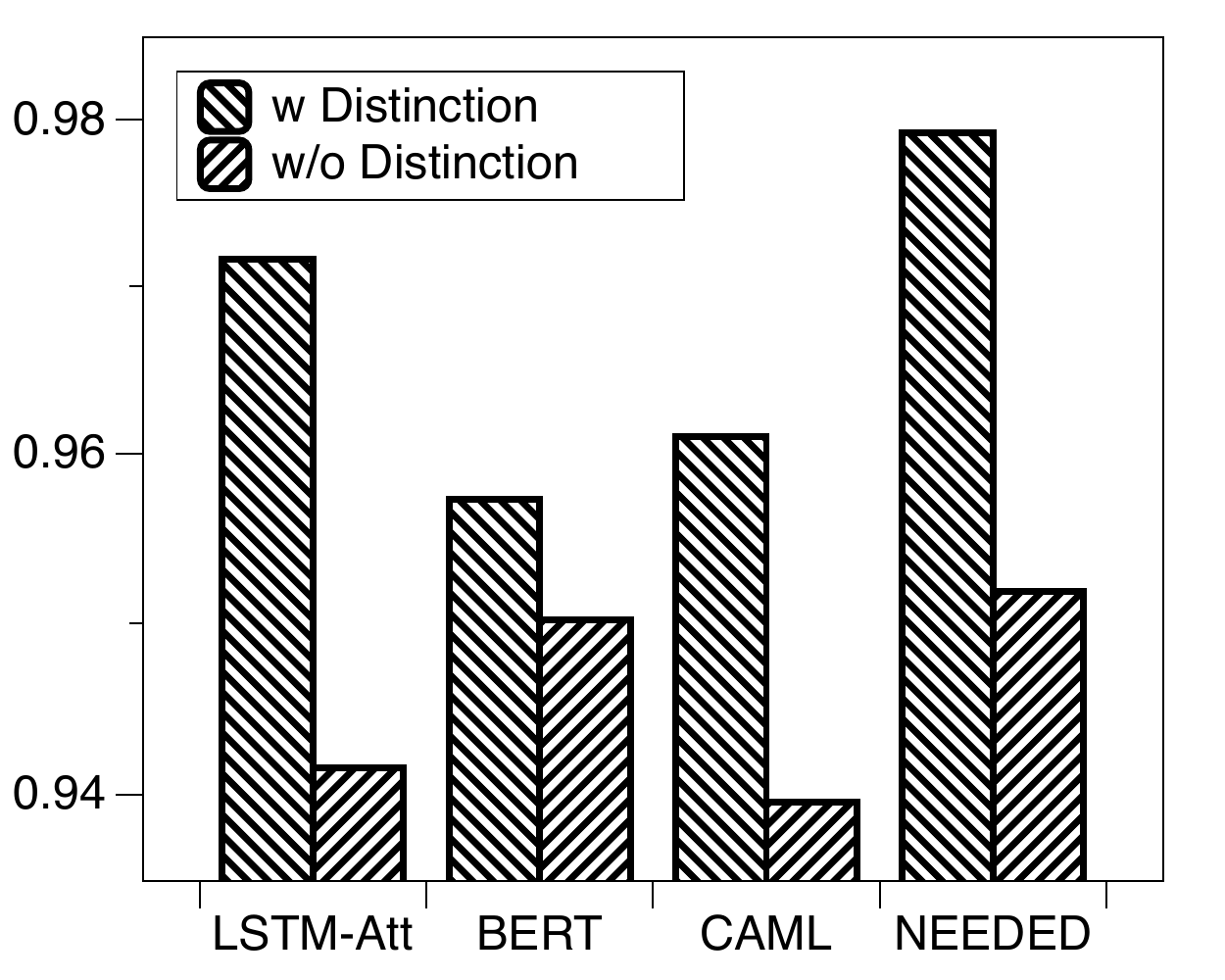}}
		\vspace{-0.2cm}
	\caption{The effect of distinguishing descriptions relevant to different eyes in the preprocessing step.}
		\label{fig_distinction}
		\vspace{-0.5cm}
\end{figure*}


\subsubsection{RQ1: Overall Comparison}
The goal of the first experiment is to compare the performance of the NEEDED and baseline models on the ADED task. In this comparison, we used Macro-F1, Micro-F1, Macro-AUC, and Micro-AUC as evaluation metrics, which are widely used in multi-label classification problems~\cite{xiao2021expert, CAML,zhou2021automatic, chen2021novel}. The results are shown in Table \ref{experiment}.
From the overall results, we observed that: First, our framework achieves the best performance in overall evaluation metrics, showing it is better at extracting disease-related information from the OEMR document. Second, our model performs better than w/o c by learning both word-level and sentence-level dependencies and preserving text type information. 
Third, BERT as pre-trained model does not perform as well as other baseline models on most metrics, probably due to the large gap between the text in its pre-training corpus and the OEMR documents.

\begin{figure}[!htbp]
	\centering
	\subfloat[MiF1 \& MaF1]{
		\includegraphics[width=4.0cm]{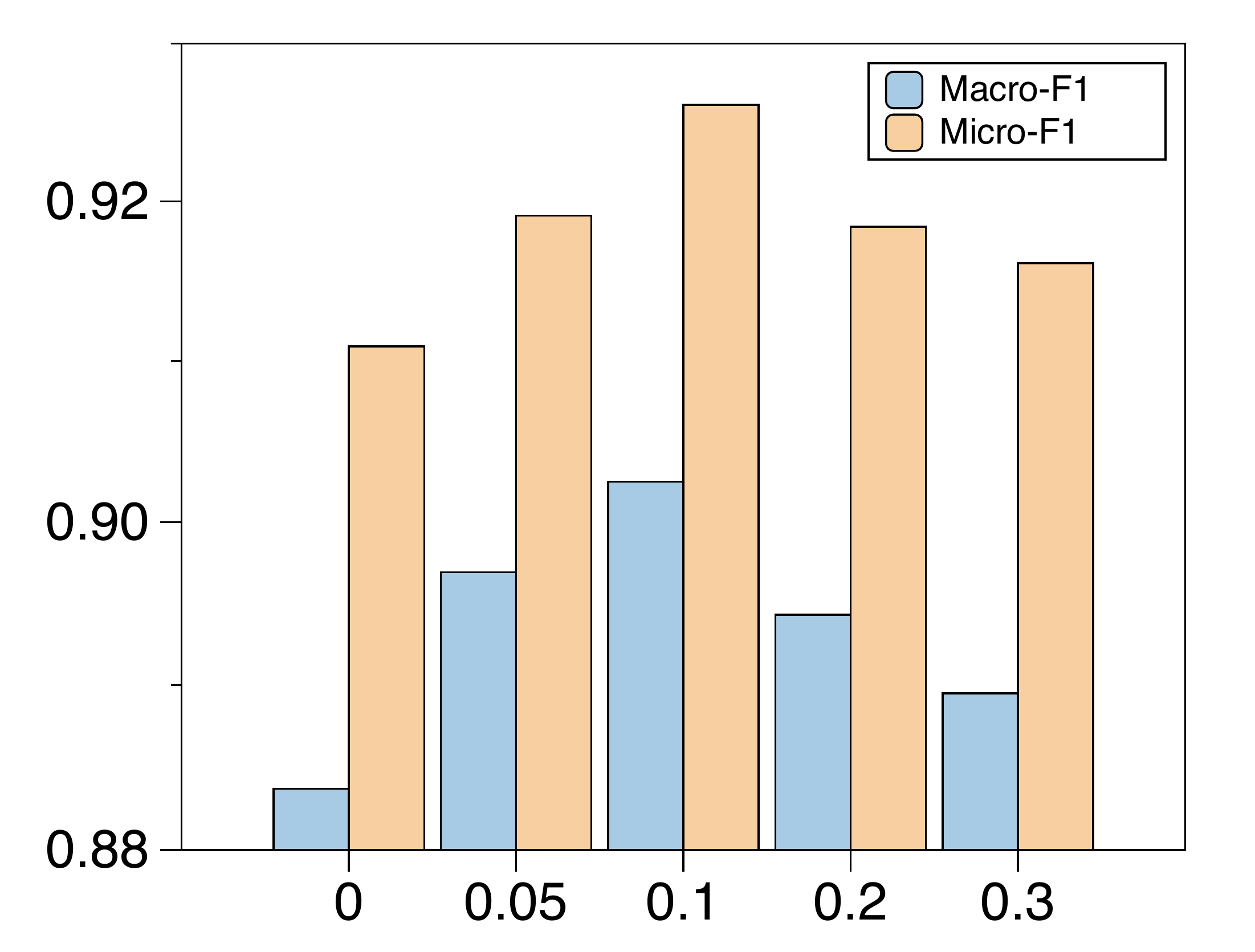}}
		\hspace{-3mm}
	\subfloat[MiAUC \& MaAUC]{
		\includegraphics[width=4.0cm]{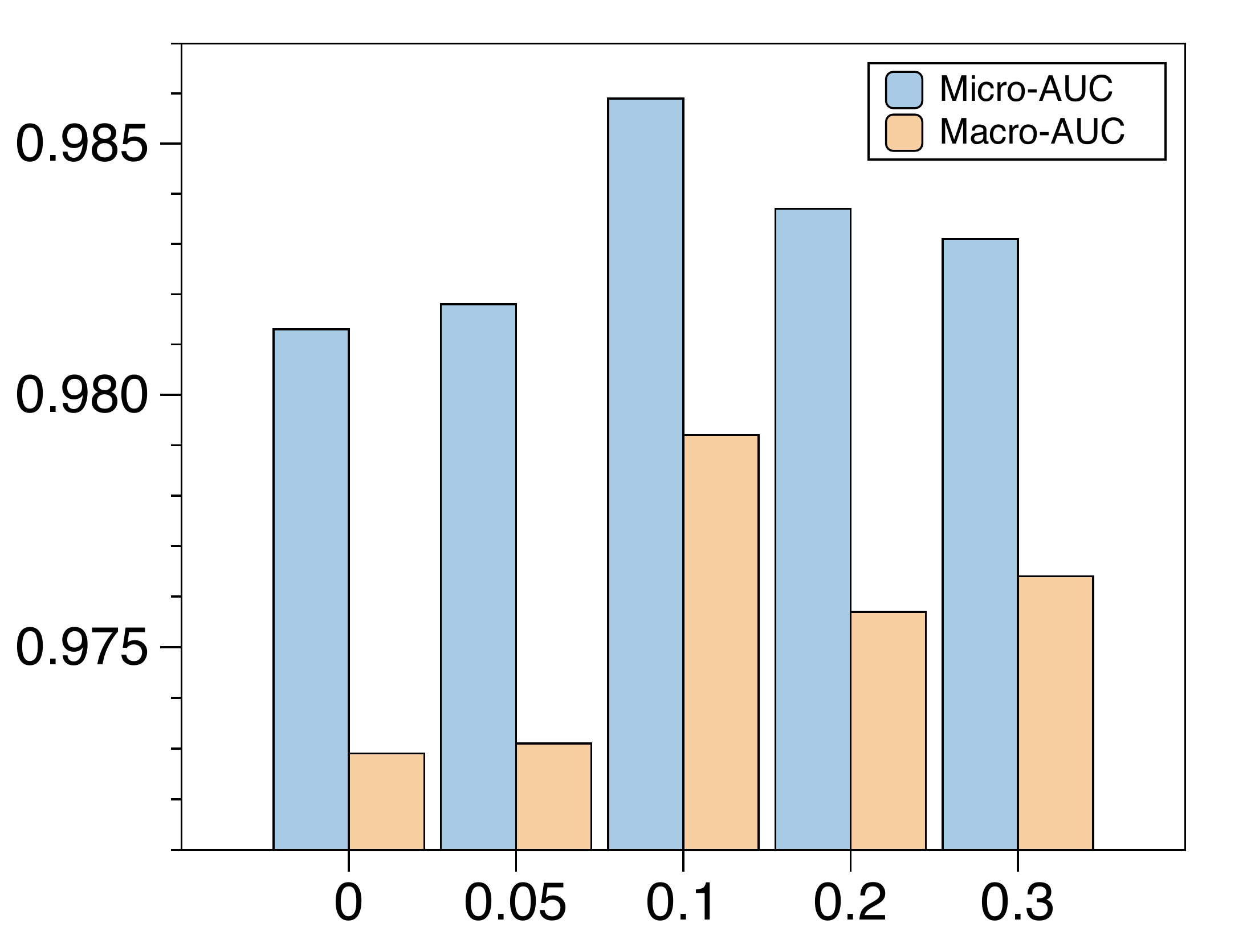}}
		\hspace{-3mm}
		\vspace{-0.2cm}
	\caption{NEEDED performance in different templated asymptomatic description filtering thresholds, where `0' represent the ablation of the filter.}
		\label{filter}
		\vspace{-0.5cm}
\end{figure}

\subsubsection{RQ2: Ablation Study of NEEDED} To explore the effectiveness of each part of our framework, we conduct ablation experiments. 
We observed that our framework outperforms all its variants, illustrating that each component in our framework is indispensable. In addition to this, there are the following points:
First, compared to our model, both w/o c and w/o w show a significant performance drop due to the fact that they both remove the hierarchical Transformer. Meanwhile, w/o c is better than w/o w in most metrics because it retains the attention-based predictor.
Second, w/o s outperforms w/o w in all metrics and w/o c is better than w/o w in most metrics, indicating the effectiveness of attention-based predictor which can capture disease-specific information. 
Third, although the performance of w/o l is degraded compared to our framework, it performs better than w/o w in all metrics. This is because it removes the attention-based predictor while retaining the hierarchical Transformer, indicating the importance of learning the dependency between sentences.
Fourth, w/o p is the worst and has a very significant degradation compared to the other models because it removes the preprocessing component, resulting in clutter and sparsity of information in the model input.

\subsubsection{RQ3: The Impact of Distinguishing Descriptions Relevant to Different Eyes}
To explore the effect of distinguishing descriptions of different eyes, we conduct the following experiment. We first filter part of the asymptomatic descriptions in the OEMR document according to the previous section. Then, instead of distinguishing the remaining descriptions to form two documents, we use all filtered texts as model inputs to obtain the diagnosis results for both eyes. We perform this experiment on our framework and all baseline methods, then compare the results with the case where distinction were made. The results are shown in figure \ref{fig_distinction}. 
We observe that the performance of all models degrades significantly if the descriptions are not distinguished. The reason is the information clutter caused by descriptions of different eyes mixed up. Besides, we observed that our model is the most affected, probably because it utilize the hierarchical Transformer to model the document's information at sentence level, which exacerbate the information clutter.
In addition to the above, we observed that BERT is the least affected, probably because it has a huge number of model parameters and is pre-trained on a large corpus, making its fitting ability more stable despite the information clutter.


\begin{figure*}[!ht]
\centering
\subfloat[Hyperparameter study on transformer layer number.]{  
	\includegraphics[width=0.33\linewidth]{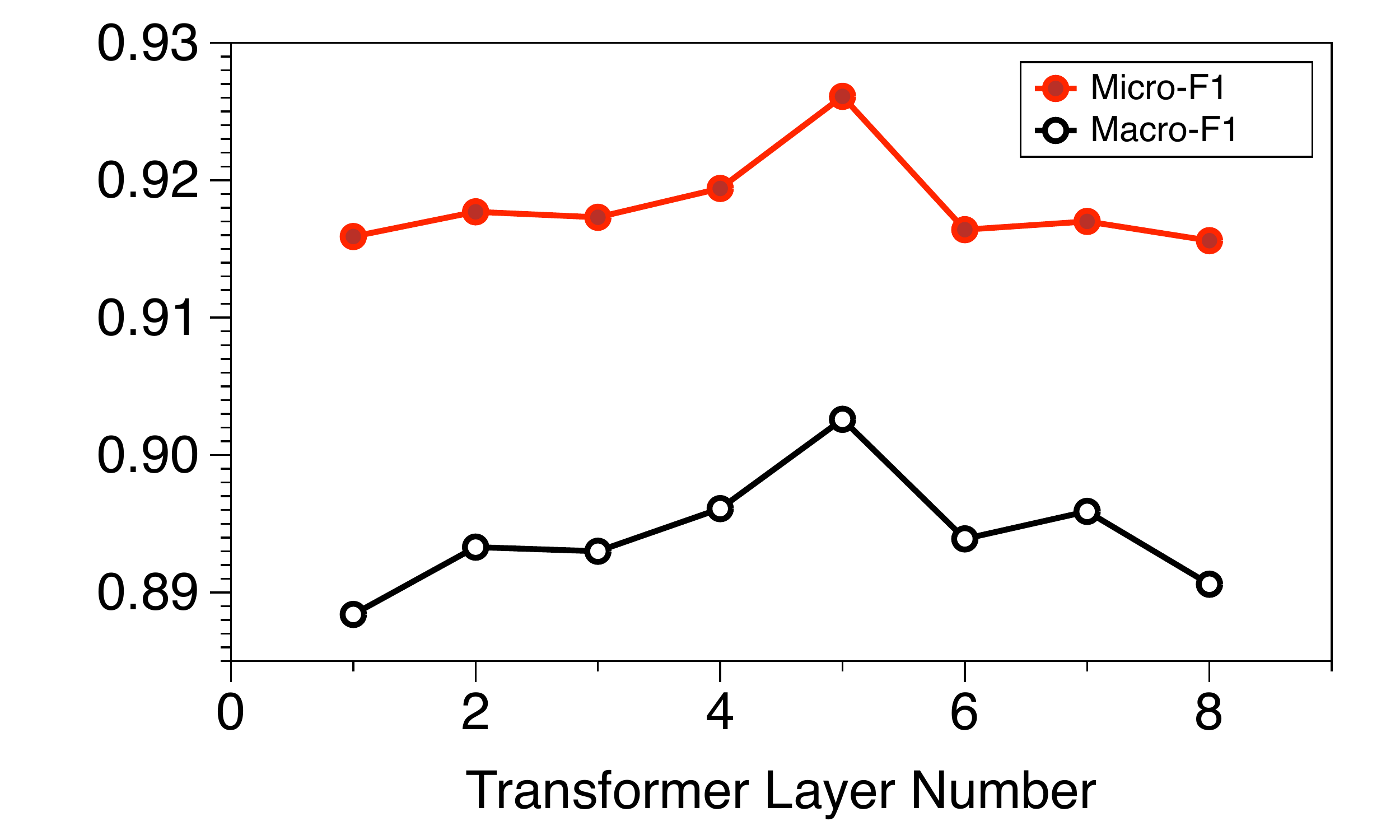}}
	\hspace{-3mm}
\subfloat[Hyperparameter study on attention head number.]{
	\includegraphics[width=0.33\linewidth]{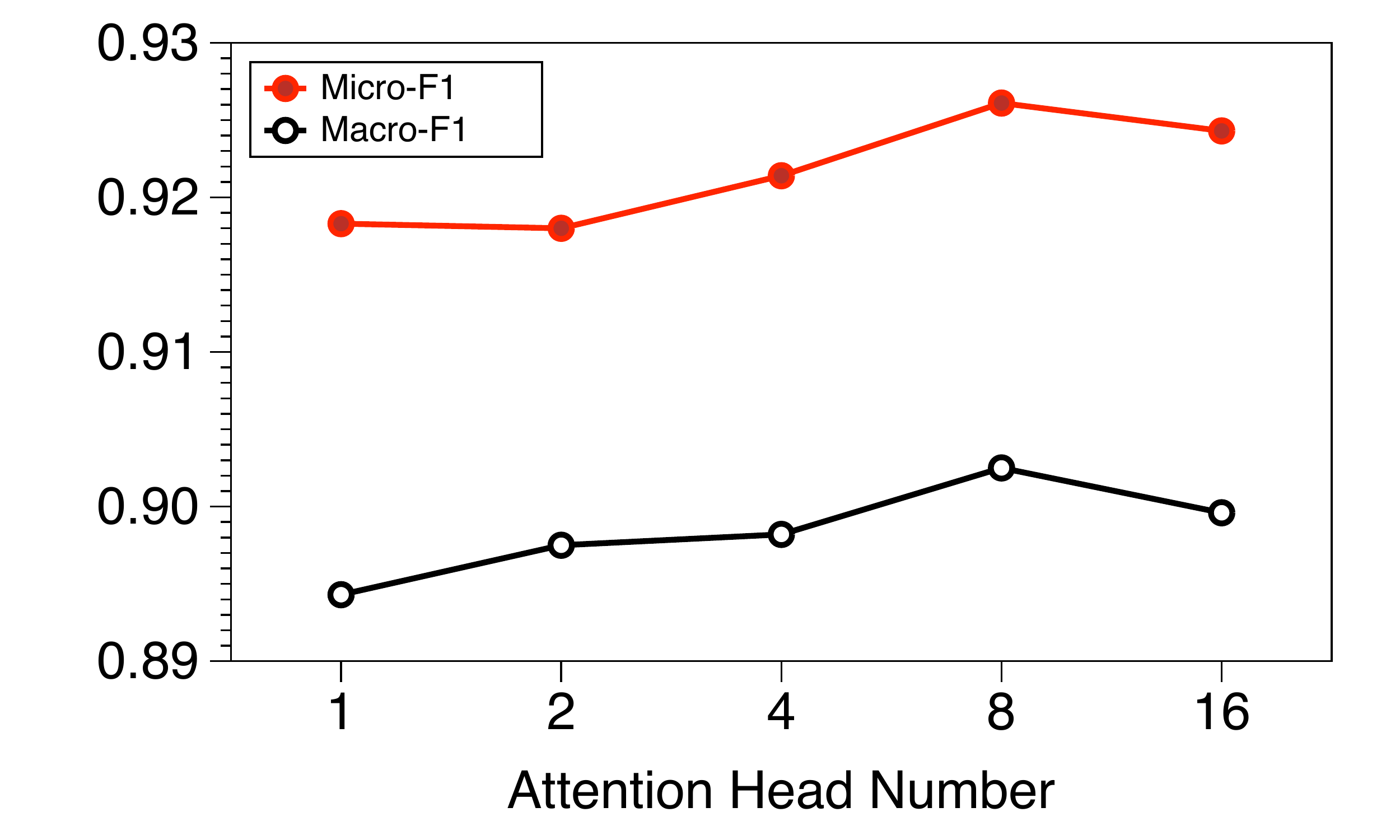}}
	\hspace{-3mm}
\subfloat[Model convegence study]{
	\includegraphics[width=0.33\linewidth]{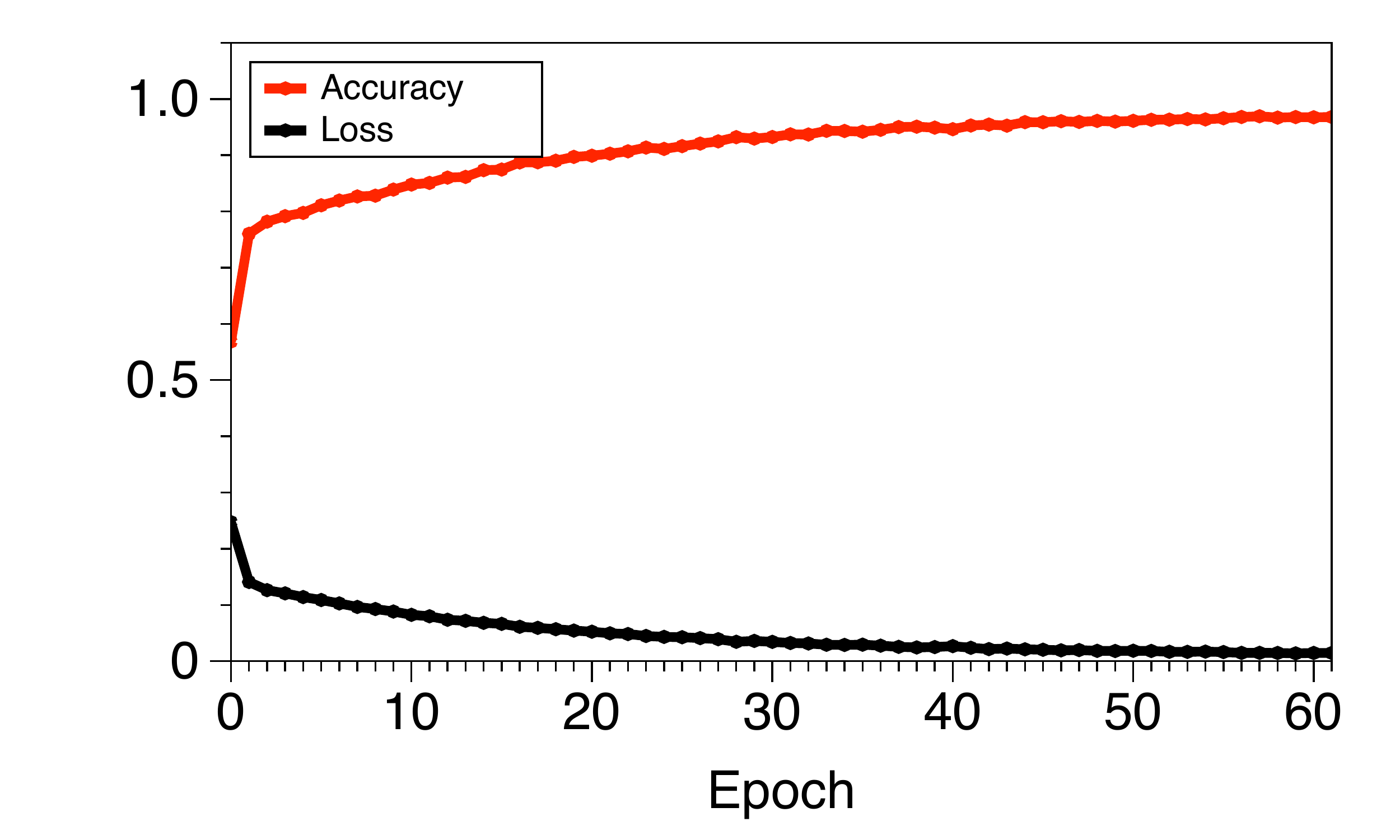}}
\caption{Hyperparameter study and model convergence study.}
	\label{fig_hyper}
	\vspace{-0.5cm}
\end{figure*}

\subsubsection{RQ4: The Impact of Filtering Templated  Asymptomatic Descriptions}
In our framework, 
we filter out templated asymptomatic descriptions that occur more frequently than a certain threshold before the OEMR document is fed into the downstream model.
In this experiment, we explore the impact of this way by setting different filtering thresholds and not filtering at all. Specifically, we set the filtering thresholds $B$ to $0.05$, $0.1$, $0.2$, and $0.3$, as well as no filtering, to observe the changes in the performance of our framework.
The results are shown in figure \ref{filter}, where $0$ represent the ablation of filter.
We can find that after the threshold $B$ is bigger than $0.1$, the model's performance tends to decrease as $B$ increases. The performance is worst in the absence of filtering, which indicates the validity of filtering the templated asymptomatic description.

We also observe that the performance is worse when $B$ is $0.05$ than when $B$ is $0.1$, indicating that the too-low filtering threshold is harmful. This may be due to some valuable asymptomatic descriptions being filtered out.


\subsubsection{RQ5: Exploring the Explainability of NEEDED}
\begin{figure}[!htbp]
\centering
\includegraphics[width=0.48\textwidth]{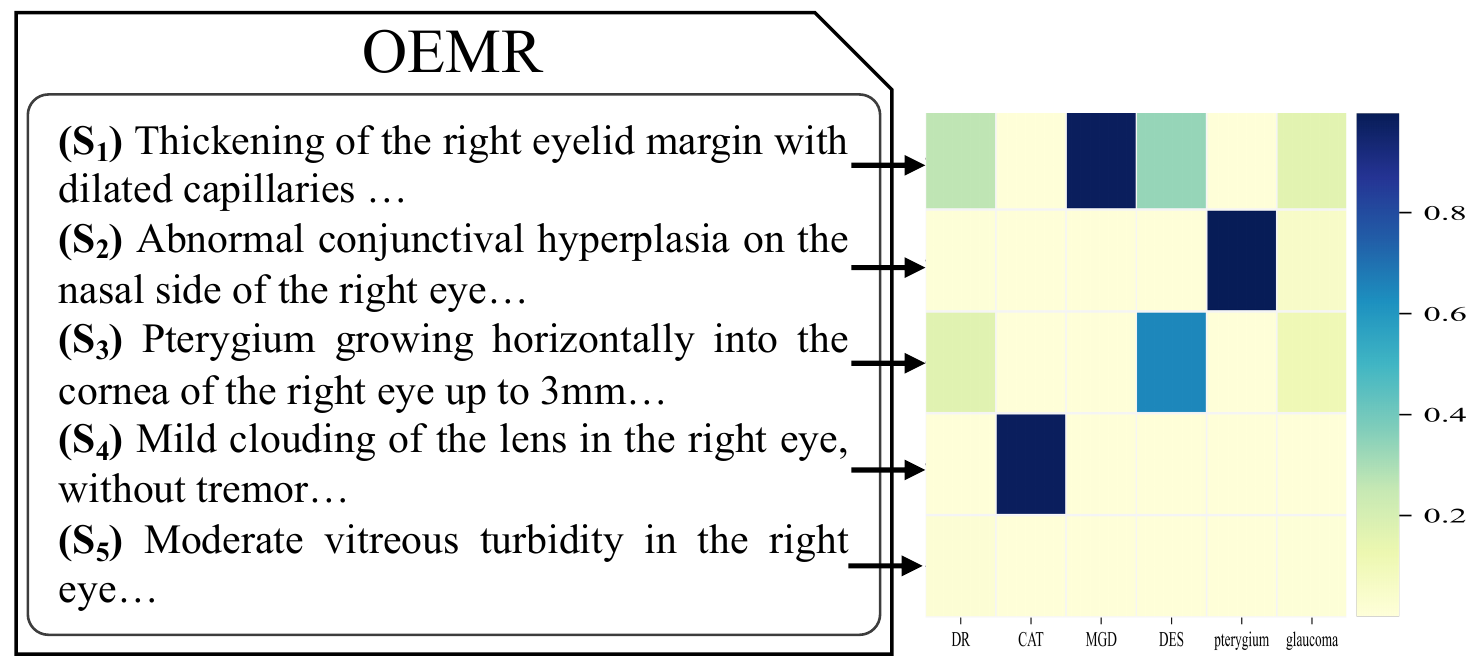}
\caption{Explainability Case Study. The left part of this figure shows a portion of an OEMR document describing the patient's right eye, and the right part shows a heatmap in which each column is the distribution of attention weights for one disease.}
\label{fig:expla}
\vspace{-0.5cm}
\end{figure}

Our predictor utilizes dot-product attention to extract disease-specific information, which improves the model's effectiveness and can provide some explainability to the diagnostic results.
Specifically, assume that our model diagnoses the disease $l$ based on the given input, we can obtain the attention weight vector $\alpha_l$ from the predictor. The $\alpha_l^i$ denotes the attention weight of the $i$-th sentence's representation for the disease $l$, which can be viewed as the importance of the $i$-th sentence for diagnosis of disease $l$. Thus, for each disease, we can obtain the important sentences associated with it in the input based on the attention weights.
We show a case to demonstrate the effect of providing explainability in this way.
The left part of Figure \ref{fig:expla} shows a portion of an OEMR document describing the patient's right eye.
Specifically, the patient's right eye was diagnosed with cataract(CAT), meibomian gland dysfunction(MGD), dry eye syndrome(DES), and pterygium by doctor. Our model yields the same diagnosis. 
The right part of Figure \ref{fig:expla} shows a heatmap in which each column is the distribution of attention weights for one disease.
We can observe that: First, our model can focus on different information and make the correct diagnosis when diagnosing various diseases, demonstrating that our model can capture disease-specific information from the document by the attention mechanism. Second, based on the distribution of attention weights, we can track the sentences essential for diagnosing the disease, which provides explainability to our model. Third, we can learn each sentence's role in diagnosing different diseases based on the distribution of attention weights, which can informing clinical research on the condition.

\vspace{-0.4cm}

\subsubsection{Hyperparameter Sensitivity and Model Converge Study}
The number of Transformer layers in the word-level Transformer and the number of self-attention heads in the hierarchical Transformer are two critical parameters in our model. To assess the impact of these two hyperparameters, we investigate the sensitivity of them and show the results in Figure \ref{fig_hyper}. We observe that the performance of our model shows a trend of increasing and then decreasing as the number of Transformer layers varies from one to eight. In addition, the performance of our model shows the same trend as the number of self-focused heads increases. This may be because the fitting ability of the model improves with more parameters introduced, but after introducing too many parameters, the model appears to be over-fitted.
Besides, we collected the loss values and accuracy of each epoch during the training process and visualized them to study the convergence of our model. We observe that the model's accuracy increases rapidly in the first ten training epochs, then tends to converge. The loss curve shows that the model's training process is stable.




\section{RELATED WORK}
\subsection{Clinical Text Classification}
With the development of NLP techniques, various neural models have been proposed to address the problem of text-based automatic disease diagnosis and other clinical text classification tasks.

Many studies treat medical text as free text and propose methods based on neural networks\cite{sha2017rnn,schoene2019rnn,yang2018cnn,girardi2018cnn,hashirMortality,yuwono-etal-2019-learning} to solve the medical text classification problem. 
For instance, Yang et al.\cite{yang2018cnn} train a multi-layer CNN to predict several basic diseases based on electronic medical records.
Girardi et al.\cite{girardi2018cnn} propose an attention-based CNN model to assess patient risk and detect warning symptoms. 
Yuwono et al.\cite{yuwono-etal-2019-learning} combine CNN and RNN to diagnose appendicitis based on clinical notes.
Hashir and Sawhney \cite{hashirMortality} established a hierarchical neural network to predict the mortality of inpatients using clinical records as input.
Some work extracts structured knowledge from clinical text and then uses it to improve model performance or get results directly~\cite{chen2020towards,yuan2021graph}. For example, Chen et al.\cite{chen2020towards} propose a disease diagnosis framework based on CNN and Bayesian networks, which extracts symptom entities from medical records and then uses them to enhance model performance. Yuan et al.\cite{yuan2021graph} first extract medical entities from medical records and construct a graph based on the relationships between them, and later extract structured information from the graph for disease diagnosis.

Nowadays, transformers~\cite{vaswani2017attention} are widely adopted in various scenarios~\cite{xiao2021expert, xiao2022imbalance,feng2020explainable,si2021hierarchical}. 
For clinical text classification, Feng and Shaib~\cite{feng2020explainable} build a hierarchical CNN-transformer model for sepsis and mortality prediction. Si and Roberts~\cite{si2021hierarchical} propose a clinical document classification model based only on transformers.

\subsection{Automatic Diagnosis of Eye Diseases}
Many scholars have explored how to automatically diagnose eye diseases based on artificial intelligence. Most of these studies are based on image data~\cite{christopher2018performance,keel2019visualizing,milea2020artificial,octlee2017deep}. For example, Christopher et al. ~\cite{christopher2018performance}train several neural networks with ocular fundus photographs as input to diagnose glaucomatous optic nerve damage. Stuart Keel et al.~\cite{keel2019visualizing} build two CNN neural networks based on fundus images to determine the presence of diabetic retinopathy and glaucoma, respectively. 
Besides, there are a few studies based on text data~\cite{apostolova2017opentext,saleh2018learningtext}. Apostolova et al.~\cite{apostolova2017opentext} built an open eye injury identification model based on TF-IDF and SVM. Saleh et al.~\cite{saleh2018learningtext} extracted numerical type features from the electronic health records of diabetic patients and used machine learning models to predict the risk of diabetic retinopathy.

\section{Conclusion}
This paper presents a framework for the automatic diagnosis of eye diseases based on OEMR, which consists of three modules: a preprocessing component, a hierarchical transformer encoder, and an attention-based predictor.
In the processing step, we filter out the useless descriptions and then distinguish the content relevant to different eyes in the OEMR document to form two documents.
Then, we obtain the representation matrix for the input document based on the hierarchical transformer encoder. Finally, we utilize dot-product attention to capture the disease-specific information and make the diagnosis. 
Experiments on the real dataset show the superiority and explainability of our framework, and each component in our framework is effective.

\section{Ackownledgement}
This work was supported by the National Key Research and Development Plan of China under Grant No. 2022YFF0712200
,the Natural Science Foundation of China under Grant No. 61836013
,Science and Technology Service Network Initiative, Chinese Academy of Sciences under Grant No. KFJ-STS-QYZD-2021-11-001
,Beijing Natural Science Foundation under Grant No. 4212030
,Youth Innovation Promotion Association CAS

\bibliographystyle{unsrt}
\bibliography{ref}
\end{document}